\RequirePackage[svgnames,table]{xcolor}

\documentclass[11pt,letterpaper]{mystyle}

\usepackage[all]{hypcap}
\usepackage[comma,numbers,compress]{natbib}
\usepackage{multirow}
\usepackage{adjustbox}
\usepackage{wrapfig}
\usepackage{placeins}
\usepackage{xspace}
\usepackage{array}
\usepackage{nicefrac}
\usepackage{xurl}
\usepackage{hyperref}
\usepackage{graphicx}

\tcbuselibrary{skins,breakable}

\definecolor{eeveebg}{RGB}{220, 235, 248}
\definecolor{gaincolor}{RGB}{35, 135, 78}
\definecolor{dropcolor}{RGB}{188, 62, 55}
\definecolor{casegreenbg}{RGB}{236, 248, 241}
\definecolor{casegreenborder}{RGB}{80, 145, 104}
\definecolor{casebluebg}{RGB}{235, 243, 252}
\definecolor{caseblueborder}{RGB}{78, 117, 164}
\definecolor{caseorangebg}{RGB}{253, 244, 232}
\definecolor{caseorangeborder}{RGB}{177, 115, 55}
\definecolor{darkblue}{rgb}{0.3, 0.5, 0.9}
\definecolor{EeveeTitleBlue}{RGB}{31, 78, 121}
\definecolor{EeveeHeaderBlue}{RGB}{26, 82, 139}
\definecolor{EeveeBoxBlue}{RGB}{232, 242, 252}
\definecolor{EeveeBoxBorder}{RGB}{210, 227, 244}

\hypersetup{citecolor=darkblue, linkcolor=darkblue, urlcolor=darkblue}
\renewcommand{\titlefont}{\centering\color{EeveeTitleBlue}\normalfont\fontsize{21}{23}\selectfont}
\renewcommand{\headingfont}{\color{EeveeHeaderBlue}\bfseries\fontsize{13}{14}\selectfont}
\tcbset{
  titlebox/.style={
    enhanced,
    colback=EeveeBoxBlue,
    colframe=EeveeBoxBorder,
    boxrule=0.5mm,
    arc=2mm,
    auto outer arc,
    left=5mm,
    right=5mm,
    top=5mm,
    bottom=5mm,
  }
}

\newcommand{\posdelta}[1]{\hspace{0.15ex}\raisebox{-0.45ex}{\scriptsize\textcolor{gaincolor}{+#1}}}
\newcommand{\negdelta}[1]{\hspace{0.15ex}\raisebox{-0.45ex}{\scriptsize\textcolor{dropcolor}{-#1}}}

\newcommand{\projectpageurl}{https://princeton-ai2-lab.github.io/EEVEE/}
\newcommand{\codeurl}{https://github.com/Princeton-AI2-Lab/EEVEE}
\newcommand{\websitelogo}{\raisebox{-1.5pt}{\includegraphics[height=1.05em]{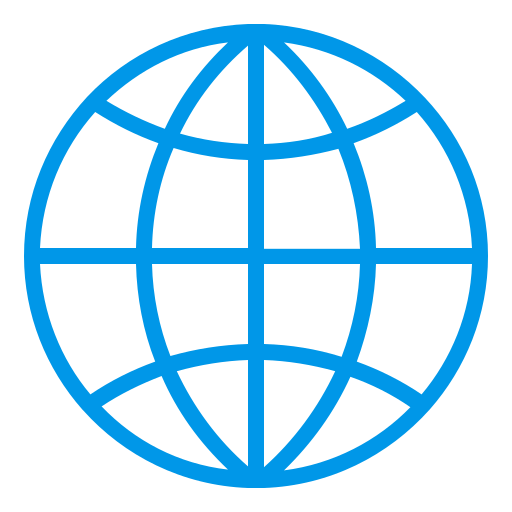}}}
\newcommand{\githublogo}{\raisebox{-1.5pt}{\includegraphics[height=1.05em]{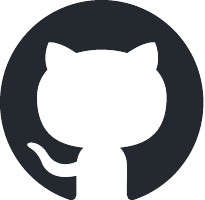}}}
\newcommand{\iconsize}{1em}
\newcommand{\icosjtu}{\raisebox{-2pt}{\includegraphics[height=1.25\iconsize]{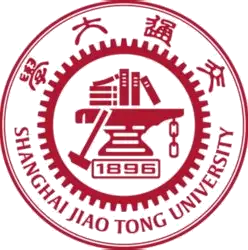}}}
\newcommand{\icoprinceton}{\raisebox{-2pt}{\includegraphics[height=1.25\iconsize]{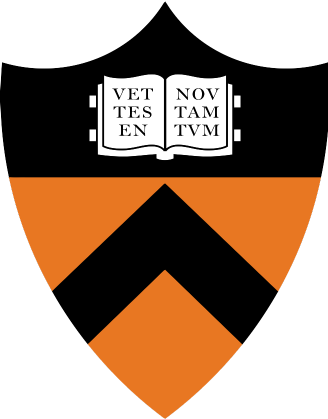}}}
\newcommand{\projectlinks}{%
  {\normalfont\small
  \websitelogo\,\href{\projectpageurl}{\textbf{Website}} \quad
  \githublogo\,\href{\codeurl}{\textbf{Code}}}%
}
\renewcommand{\titleboxlogo}{\includegraphics[width=3.15cm]{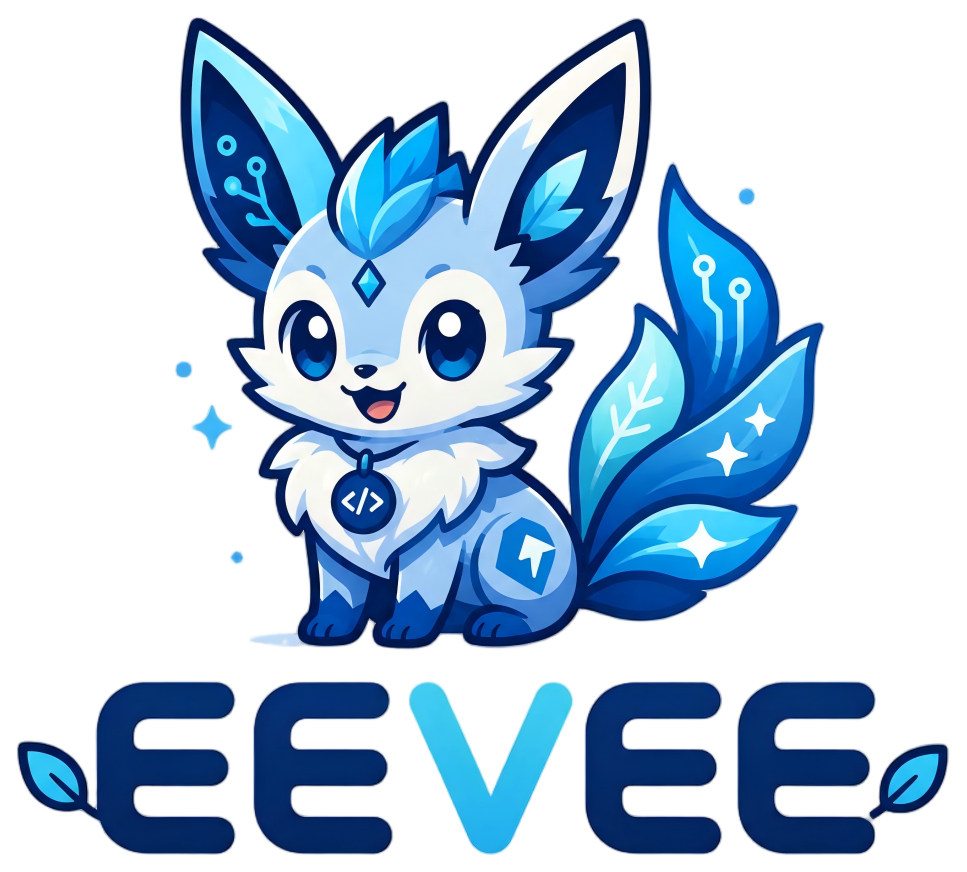}}

\makeatletter
\renewcommand{\maketitle}
{
\vphantom{a}
\vspace{-10mm}
\begin{tcolorbox}[titlebox,
  overlay={
    \node[anchor=south east, inner sep=0pt] at ([xshift=-1.50cm,yshift=-0.90cm]frame.north east) {\titleboxlogo};
  }]
\bgroup\setlength{\parindent}{0pt}
  \vskip14pt
  {\centering \titlefont \@title\par}
  \vskip11pt
  {\centering \@author\par}
  \vskip13pt
\egroup
{%
  {\abscontent}
}%
\thispagestyle{firststyle}
\end{tcolorbox}
}
\makeatother

\def\Snospace~{Section }

\title{\textsc{Eevee}: Towards Test-time Prompt Learning in the Real World for Self-Improving Agents}
\runningtitle{\textsc{Eevee}: Towards Test-time Prompt Learning in the Real World for Self-Improving Agents}

\author{
  Weixian Xu$^{1,2,*}$,
  Shilong Liu$^{2,\dagger}$, and
  Mengdi Wang$^{2,\dagger}$
  \protect\\[2mm]
  {\normalfont\small
  $^1$\icosjtu\,Shanghai Jiao Tong University \quad
  $^2$\icoprinceton\,Princeton University}
  \protect\\[2mm]
  \projectlinks
}

\date{}

\begin{document}

\begin{abstract} 
  In this paper, we propose \textsc{Eevee}, the first {\textit{multi-dataset test-time prompt learning}} framework for LLM agents, enabling test-time prompt learning under real-world task streams. 
  Existing methods are largely designed for single-dataset settings, while real-world applications require models to handle heterogeneous input streams drawn from multiple datasets, domains, and task distributions, limiting their practical applicability.
  To mitigate cross-dataset interference, \textsc{Eevee} introduces a \textit{router} that partitions incoming inputs into task clusters and assigns them to suitable prompt configurations. This design is optimized via a \textit{router-prompt co-evolution strategy}, which employs interleaved router and prompt learning phases to address their mutual dependency.
  Experiments across multiple datasets demonstrate that the framework improves robustness under heterogeneous data streams while maintaining single-benchmark learning capability and efficiency.
  Specifically, \textsc{Eevee} improves average multi-benchmark scores by 10.38 and 24.32 points over Qwen3-4B-Instruct and DeepSeek-V3.2, surpassing SOTA methods GEPA and ACE by up to 37.2\% and 48.2\%.
  
\end{abstract}

\maketitle
\correspondingauthor{$^{*}$\textit{First author} \quad $^{\dagger}$\textit{Corresponding authors} \quad \textit{Work done while Weixian Xu interned at Princeton AI Lab.}}

\section{Introduction}

\begin{figure}[!h]
    \centering
    \includegraphics[width=0.96\linewidth]{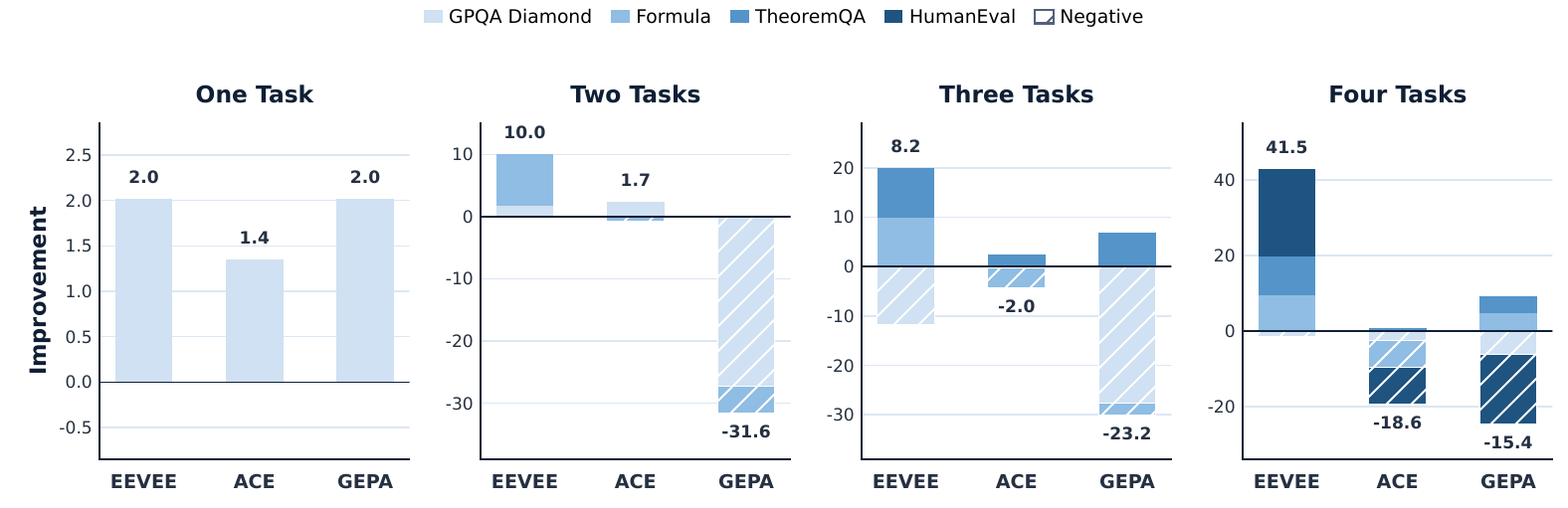}
    \caption{Incremental multi-benchmark retention improvement as tasks are added in the order GPQA Diamond, Formula, TheoremQA, and HumanEval. Each bar stacks per-benchmark improvements for all tasks seen so far: solid upward blocks are positive gains, and hatched downward blocks are negative retention losses. The number above or below each bar is its final summed improvement after all blocks are added.}
    \label{fig:multi-benchmark-curve}
\end{figure}

\textit{Test-time prompt learning} offers a lightweight mechanism for adapting foundation models after deployment.
Prior work shows that prompts can serve as an effective adaptation interface without updating model weights~\cite{lester2021power}, and black-box optimizers can revise instructions from model feedback~\cite{pryzant2023automatic,yang2024large}.
Rather than relying on a fixed offline prompt, test-time prompt learning updates prompts for new inputs, distribution shifts, and failure modes.
This makes it suitable for self-improving agents, where behavior is refined through interaction with the environment, as in reflective agents and evolving-context systems~\cite{shinn2023reflexion,zhang2026agentic}.

Recent methods such as GEPA~\cite{agrawal2026gepa}, ACE~\cite{zhang2026agentic}, and Combee~\cite{li2026combee} improve test-time prompt learning through reflection, context evolution, or scalable trace aggregation.
However, they mostly adapt within a single dataset or benchmark.
In real-world deployment, incoming queries often come from heterogeneous domains, task formats, and capability mixtures.
We formalize this regime as \textit{multi-dataset test-time prompt learning}: a model receives a stream of examples drawn from multiple datasets and domains rather than one stationary source.

This setting exposes cross-dataset interference.
Existing methods often assume a unified adaptation objective, a fixed prompt space, or feedback from one benchmark, so updates for one domain can harm another.
As Figure~\ref{fig:multi-benchmark-curve} shows, when more benchmarks enter the adaptation stream, GEPA~\cite{agrawal2026gepa} and ACE~\cite{zhang2026agentic} accumulate negative retention on previous tasks, suggesting that a single learned prompt struggles to absorb heterogeneous feedback without losing task-specific behavior.
This motivates a framework that can preserve specialization while still learning from a mixed stream of tasks.

We propose \textsc{Eevee}, a test-time prompt learning framework that augments prompt learning with a \textit{router}.
Instead of forcing all inputs through one adaptation path, the router partitions the stream into task clusters and assigns each cluster to a suitable prompt configuration.
This preserves prompt-based adaptation while reducing destructive interference across domains.

Designing the router, however, is itself difficult.
A rigid router fails to capture diverse task structure, while an unstable router disrupts prompt optimization.
The prompt learner and router are also coupled: routing determines which examples each prompt learns from, and prompt behavior determines which routing policy is useful.
We therefore introduce a \textit{router-prompt co-evolution strategy} that interleaves router and prompt learning phases, allowing routing decisions and prompt updates to improve together rather than being fixed or trained in isolation. To make this co-evolution practical, we further design a three-stage training process that initializes useful prompt slots, explores coupled updates efficiently, and then converges under a stable router.

We evaluate \textsc{Eevee} on multiple datasets and show that it consistently outperforms competitive baselines for multi-dataset test-time prompt learning.
Across the four-benchmark suite, \textsc{Eevee} improves the average score by 10.38 and 24.32 points over Qwen3-4B-Instruct~\cite{yang2025qwen3} and DeepSeek-V3.2~\cite{deepseekai2025v32}, and by up to 37.2\% and 48.2\% over GEPA and ACE.
In the incremental multi-benchmark setting, \textsc{Eevee} ends with a +41.53 cumulative retention gain after all tasks are introduced, while GEPA and ACE end at -15.36 and -18.58.
Single-benchmark and token-cost analyses further show that the routing design remains competitive in conventional single-task settings while avoiding ACE's large prompt expansion.

Contributions:
\begin{itemize}
    \item We propose \textsc{Eevee}, the first \textit{multi-dataset test-time prompt learning} framework for LLM agents, using a router to reduce cross-dataset interference.
    \item We introduce \textit{router-prompt co-evolution}, enabled by a three-stage training design, to jointly improve router prompts and model prompts through interleaved phases.
    \item We validate \textsc{Eevee} on multiple datasets, showing strong performance, retention, and efficiency, with case studies that offer practical guidance.
\end{itemize}

\section{Methods}

\subsection{Framework Overview}
\label{sec:framework-overview}

\textsc{Eevee} targets multi-dataset test-time prompt learning, where a mixed stream contains different domains, formats, and evaluation rules. A single evolving prompt can let updates for one task family interfere with another, so \textsc{Eevee} maintains a set of specialized prompts and a router that chooses which prompt should handle each input. We denote the prompt set by $\mathcal{P}=\{p_1,\ldots,p_K\}$ and the router by $R$. The target model $M$ is fixed.

At inference time, the router first selects a slot and the target model then answers with the corresponding prompt:
\[
    z = R(x;\mathcal{P}) \in \{1,\ldots,K\},
    \qquad
    \hat{y}=M(x;p_z).
\]
This preserves prompt-based adaptation while allowing different inputs to invoke different specialized behaviors.

To obtain strong performance, the router itself must be learned: default or manually written routers do not reliably recover behaviorally useful partitions, as shown in Table~\ref{tab:ablation-results}. Router learning iterates the router prompt and has a mutual dependency with prompt learning: routing determines which examples each prompt sees, while prompt quality determines whether a routing decision is good. \textsc{Eevee} therefore learns both through router-prompt co-evolution.

\begin{figure}[h]
    \centering
    \includegraphics[width=0.98\linewidth]{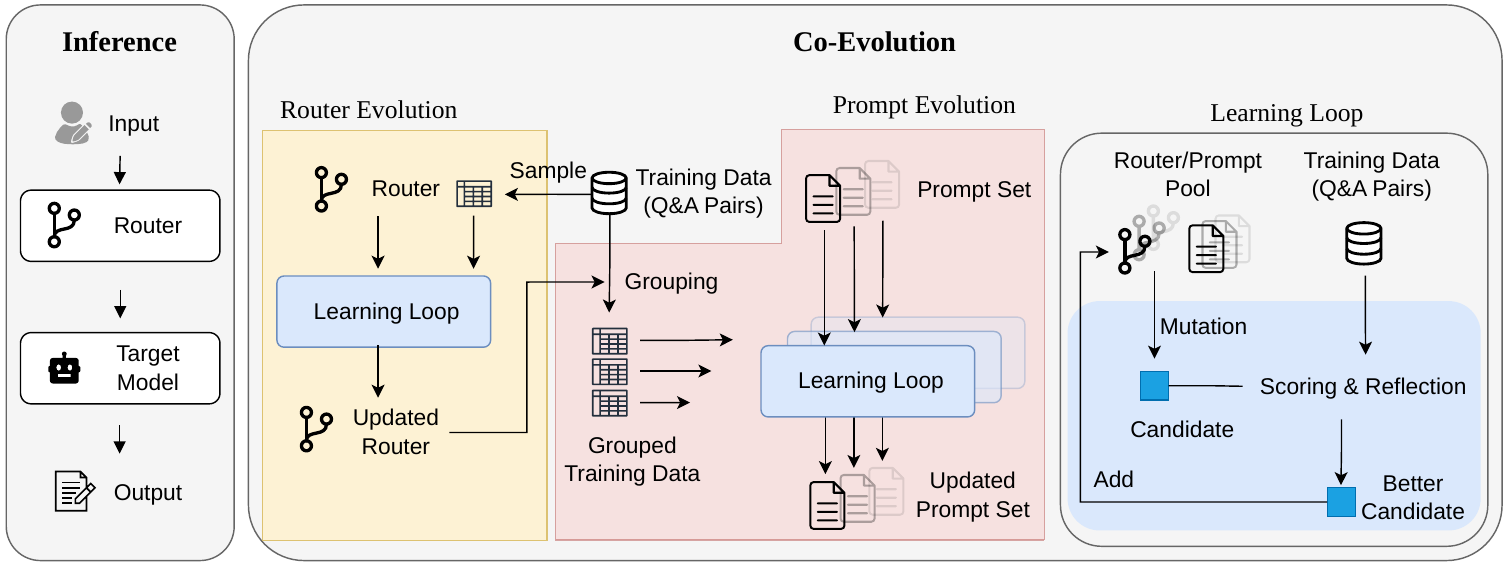}
    \caption{Main framework of \textsc{Eevee}. Inference routes each input to a specialized prompt; learning co-evolves the router and prompt set through mutation, analysis, reflection, scoring, and regrouping.}
    \label{fig:eevee-overview}
\end{figure}

\subsection{Router-Prompt Co-evolution}
\label{sec:co-evolution}

The mixed adaptation data is split into training and validation sets, $\mathcal{D}_{\mathrm{tr}}$ and $\mathcal{D}_{\mathrm{val}}$. Each co-evolution cycle alternates two operations: router evolution fixes the prompt set and searches for a better router, then prompt evolution fixes this router and updates each slot prompt on its routed data. Write $\mathcal{P}_T=\{p_{k,T}\}_{k=1}^K$. In a cycle starting from $R_T$ and $\mathcal{P}_T$,
\[
\begin{aligned}
    R_{T+1} &= \operatorname{RouterEvolve}(R_T;\mathcal{P}_T,\mathcal{D}_{\mathrm{tr}},\mathcal{D}_{\mathrm{val}}),
    & \mathcal{P}_{T+1} &= \mathcal{P}_{T},\\
    \mathcal{P}_{T+2} &=
    \left\{\operatorname{PromptEvolve}(p_{k,T+1};R_{T+1},
    \mathcal{D}_{\mathrm{tr},k}^{T+1},\mathcal{D}_{\mathrm{val},k}^{T+1})\right\}_{k=1}^{K},
    & R_{T+2} &= R_{T+1}.
\end{aligned}
\]
The next router phase uses the updated prompt set, so routing decisions and prompt specialization improve each other rather than being optimized in isolation.

\paragraph{Router evolve.}
Given $R_T$ and fixed prompts $\mathcal{P}_T$, \textsc{Eevee} initializes a temporary router pool $\mathcal{B}_R$ and repeatedly samples a router mini-batch $\mathcal{D}_{\mathrm{RM}}$. To make errors attributable to assignment rather than prompt incapability, $\mathcal{D}_{\mathrm{RM}}$ is sampled only from training examples that at least one current slot prompt can solve. Each update step samples reference routers from $\mathcal{B}_R$ and mutates them into $R_{\mathrm{mut}}$, allowing the search to redesign routing rules rather than only refine an existing router.

\textsc{Eevee} evaluates $R_{\mathrm{mut}}$ on $\mathcal{D}_{\mathrm{RM}}$. Since router quality is observed through downstream prompt correctness, \textsc{Eevee} analyzes cases where the routed slot fails but another slot succeeds, explaining why the better slot matches the task. Reflection uses these analyses and ground-truth correctness to produce $R_{\mathrm{ref}}$. Let $s_{\mathrm{mb}}(\cdot)$ denote mini-batch downstream score. \textsc{Eevee} keeps
\[
    R^\star=\arg\max_{R\in\{R_{\mathrm{mut}},R_{\mathrm{ref}}\}} s_{\mathrm{mb}}(R),
\]
evaluates $R^\star$ on $\mathcal{D}_{\mathrm{val}}$, and admits it into $\mathcal{B}_R$ only if it improves over the phase baseline $R_T$. When the router score plateaus, the phase outputs $R_{T+1}$ while keeping $\mathcal{P}_T$ fixed.

Router candidates are scored by downstream accuracy, consistency, and balance. Let $a_R(x,y)=\mathbf{1}[M(x;p_{R(x;\mathcal{P}_T)})=y]$ denote routed correctness under $R$ and the fixed prompt set:
\[
\begin{array}{rlrl}
S_R(R)&=\lambda_{\mathrm{acc}}A(R)+\lambda_{\mathrm{con}}C(R)+\lambda_{\mathrm{bal}}B(R),
&
A(R)&=\frac{1}{|\mathcal{D}_{\mathrm{val}}|}\sum_{(x,y)\in\mathcal{D}_{\mathrm{val}}}a_R(x,y),\\
C(R)&=\beta_{\mathrm{in}}\operatorname{Compact}(R)+\beta_{\mathrm{out}}\operatorname{Separate}(R),
&
B(R)&=\gamma_{\mathrm{use}}\frac{|\mathcal{K}_R|}{K}+\gamma_{\mathrm{bal}}\operatorname{Balance}(\pi_R).
\end{array}
\]
Here $\mathcal{K}_R$ is the set of labels used by $R$ and $\pi_R$ is its empirical label distribution. $\operatorname{Compact}$ rewards same-label examples with similar cached correctness vectors, while $\operatorname{Separate}$ rewards behaviorally distinguishable labels. The weights are annealed from consistency/balance toward downstream accuracy. Appendix~\ref{app:hyperparameter-robustness} shows that performance is stable across alternative annealing, weighting, and prompt-search settings.

\paragraph{Prompt evolve.}
After router evolution, $R_{T+1}$ routes the full data into slot-specific groups:
\[
\mathcal{D}_{\mathrm{tr},k}^{T+1}=\{(x,y)\in\mathcal{D}_{\mathrm{tr}}:R_{T+1}(x;\mathcal{P}_T)=k\},\quad
\mathcal{D}_{\mathrm{val},k}^{T+1}=\{(x,y)\in\mathcal{D}_{\mathrm{val}}:R_{T+1}(x;\mathcal{P}_T)=k\}.
\]
Each non-empty slot is evolved independently and in parallel. For slot $k$, \textsc{Eevee} initializes a temporary prompt pool $\mathcal{B}_{P,k}$ and samples $\mathcal{D}_{\mathrm{PM}}\subset\mathcal{D}_{\mathrm{tr},k}^{T+1}$. Prompt evolution also uses mutation and reflection, but without the router-specific analysis step: it proposes $p_{\mathrm{mut}}$ from reference prompts and mini-batch examples, then directly reflects from question, target answer, model answer, and correctness to obtain $p_{\mathrm{ref}}$. The better mini-batch prompt
\[
    p^\star=\arg\max_{p\in\{p_{\mathrm{mut}},p_{\mathrm{ref}}\}} s_{\mathrm{mb}}(p)
\]
is evaluated on the routed validation set with score $s_{\mathrm{val}}^k(\cdot)$.

\textsc{Eevee} stores prompts in a Pareto-front pool~\cite{mouret2015illuminating}. Each prompt is represented by its correctness vector over $\mathcal{D}_{\mathrm{val},k}^{T+1}$; a prompt is dominated if another is at least as correct on every example and strictly better on one. Thus the frontier preserves complementary prompts. A candidate enters $\mathcal{B}_{P,k}$ only when
\[
    s_{\mathrm{val}}^k(p^\star)>s_{\mathrm{val}}^k(p_{\emptyset}),
    \qquad
    p^\star\in\operatorname{ParetoFront}(\mathcal{B}_{P,k}\cup\{p^\star\}),
\]
where $p_{\emptyset}$ is the empty prompt. The empty-prompt floor removes ineffective edits, and the Pareto-front rule preserves diverse useful prompts. When all non-empty slots plateau, prompt evolution returns the updated prompt set.

\subsection{Training Stages}
\label{sec:training-stages}

\begin{figure}[t]
    \centering
    \begin{minipage}[c]{0.62\linewidth}
        \centering
        \includegraphics[width=\linewidth]{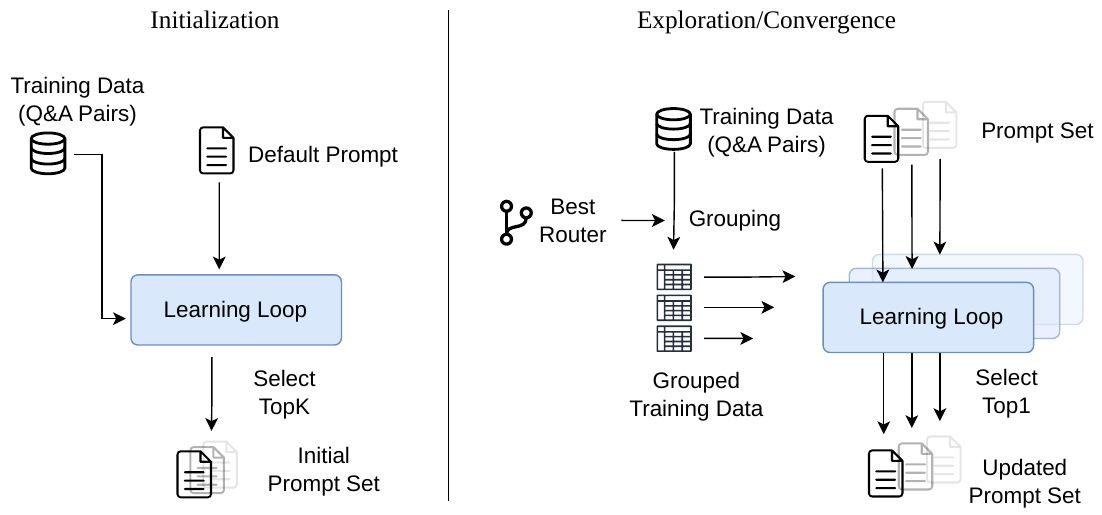}
    \end{minipage}\hfill
    \begin{minipage}[c]{0.34\linewidth}
        \centering
        \scriptsize
        \[
        \begin{array}{@{}l@{}}
        p_k=\arg\max\limits_{p\in\mathcal{B}\setminus\mathcal{S}_{k-1}}
        \Delta(p;\mathcal{S}_{k-1}),\\[0.65em]
        \Delta(p;\mathcal{S})=
        \left|C_F(p)\setminus\displaystyle\bigcup_{q\in\mathcal{S}}C_F(q)\right|,\\[0.65em]
        \mathcal{S}_{k}=\mathcal{S}_{k-1}\cup\{p_k\}.
        \end{array}
        \]
    \end{minipage}
    \caption{Three-stage training design and initialization selection rule. Left: initialization builds prompt set, exploration alternates router and prompt evolution, and convergence fixes the router for larger-budget prompt learning. Right: the greedy coverage rule for top-$K$ prompts selection.}
    \label{fig:eevee-training-stages}
\end{figure}

\textsc{Eevee} uses three stages with distinct roles: initialization creates usable prompt slots, exploration searches over coupled router-prompt designs, and convergence spends a larger budget after routing stabilizes.

\paragraph{Initialization.}
Router evolution needs a prompt set that can reveal whether a routing decision is useful; otherwise accuracy reflects prompt weakness rather than router quality. \textsc{Eevee} therefore first initializes a diverse prompt set before router learning. Initialization runs prompt learning on the mixed training set and keeps a Pareto-front pool: because frontier prompts cover complementary examples, selecting from this pool yields specialized prompts with distinguishable behavior. Let $\mathcal{B}$ be the initialization pool after dominated candidates are removed, and let $F_i$ be the reduced Pareto frontier for validation example $i$. With coverage $C_F(p)=\{i:p\in F_i\}$ and $\mathcal{S}_0=\emptyset$, \textsc{Eevee} greedily retains prompts by the rule in Figure~\ref{fig:eevee-training-stages}, where $\Delta$ denotes additional coverage. The retained prompts form $\mathcal{P}^{0}$ and provide specialized behaviors from which the initial router can be written.

\paragraph{Exploration.}
Exploration starts from $(R_0,\mathcal{P}^0)$ and alternates router and prompt evolution under lightweight budgets. Frequent switching is necessary for efficiency: fully optimizing prompts under an unstable router wastes budget, while optimizing a router against stale prompts can overfit to obsolete prompt behavior. As described in Section~\ref{sec:co-evolution}, \textsc{Eevee} enables an annealing mechanism for router-candidate scoring, shifting weights from consistency and balance toward downstream accuracy so early steps preserve diverse routing behaviors and later steps favor accurate, stable routing.

\paragraph{Convergence.}
Because exploration must switch phases frequently, it cannot fully optimize every slot prompt. Once annealing and alternating updates identify a stable router, convergence fixes $R^\star$, reroutes $\mathcal{D}_{\mathrm{tr}}$ and $\mathcal{D}_{\mathrm{val}}$, and spends a larger prompt-learning budget within each slot. This enables \textsc{Eevee} to find strong prompts under a fixed router rather than continuing to move the partition.

\section{Experiments}
\subsection{Settings}

We evaluate \textsc{Eevee} in a \textit{multi-dataset test-time prompt learning} setting over four benchmarks: GPQA Diamond~\cite{rein2023gpqa}, Formula~\cite{wang2025finlora}, TheoremQA~\cite{chen2023theoremqa}, and HumanEval~\cite{chen2021codex}. GPQA Diamond tests closed-book knowledge QA; Formula and TheoremQA emphasize mathematical and symbolic reasoning; HumanEval evaluates code generation. Unless otherwise stated, methods learn from the mixed training stream and are scored on held-out test examples outside that stream. To reduce randomness, we run stochastic settings multiple times and report averaged scores.

\subsection{Main Results}

Table~\ref{tab:main-results} compares average performance over three runs against two strong reflection-based prompt-learning baselines, GEPA~\cite{agrawal2026gepa} and ACE~\cite{zhang2026agentic}, plus the unadapted target model. On Qwen3-4B-Instruct~\cite{yang2025qwen3}, \textsc{Eevee} reaches 51.75 average score, improving over the target-model baseline by 10.38 points and outperforming GEPA and ACE by 14.02 and 16.83 points. Its gains over the baseline are +9.33 on Formula, +10.48 on TheoremQA, and +23.17 on HumanEval, where the final score reaches 72.63. On DeepSeek-V3.2~\cite{deepseekai2025v32}, \textsc{Eevee} reaches 64.07 average score, improving over the target-model baseline by 24.32 points and over GEPA by 8.24 points; the per-benchmark gains are +30.55 on Formula, +18.63 on TheoremQA, and +50.00 on HumanEval, with HumanEval reaching 92.82.

\begin{table*}[htbp]
    \centering
    \small
    \caption{Main results on the four-benchmark suite. Scores are percentages averaged over three runs. Colored subscripts denote differences from the corresponding target-model baseline.}
    \vspace{0.1cm}
    \label{tab:main-results}
    \setlength{\tabcolsep}{6pt}
    \renewcommand{\arraystretch}{1.08}
    \resizebox{0.9\textwidth}{!}{%
    \begin{tabular}{ll|cccc|c}
        \toprule
        Target model & Method & GPQA Diamond & Formula & TheoremQA & HumanEval & Avg. \\
        \midrule
        \multirow{4}{*}{\shortstack[l]{Qwen3-4B\\Instruct}}
        & Baseline & 56.00 & 45.22 & 14.79 & 49.46 & 41.37 \\
        & ACE & 48.93\negdelta{7.07} & 39.67\negdelta{5.55} & 15.84\posdelta{1.05} & 35.23\negdelta{14.23} & 34.92\negdelta{6.45} \\
        & GEPA & 50.84\negdelta{5.16} & 49.83\posdelta{4.61} & 19.62\posdelta{4.83} & 30.62\negdelta{18.84} & 37.73\negdelta{3.64} \\
        & \cellcolor{eeveebg}\textbf{\textsc{Eevee}} & \cellcolor{eeveebg}54.55\negdelta{1.45} & \cellcolor{eeveebg}54.55\posdelta{9.33} & \cellcolor{eeveebg}25.27\posdelta{10.48} & \cellcolor{eeveebg}72.63\posdelta{23.17} & \cellcolor{eeveebg}\textbf{51.75}\posdelta{10.38} \\
        \midrule
        \multirow{4}{*}{DeepSeek-V3.2}
        & Baseline & 64.98 & 30.00 & 21.21 & 42.82 & 39.75 \\
        & ACE & 55.89\negdelta{9.09} & 37.78\posdelta{7.78} & 27.05\posdelta{5.84} & 78.59\posdelta{35.77} & 49.83\posdelta{10.08} \\
        & GEPA & 41.75\negdelta{23.23} & 60.56\posdelta{30.56} & 31.72\posdelta{10.51} & 89.29\posdelta{46.47} & 55.83\posdelta{16.08} \\
        & \cellcolor{eeveebg}\textbf{\textsc{Eevee}} & \cellcolor{eeveebg}63.08\negdelta{1.90} & \cellcolor{eeveebg}60.55\posdelta{30.55} & \cellcolor{eeveebg}39.84\posdelta{18.63} & \cellcolor{eeveebg}92.82\posdelta{50.00} & \cellcolor{eeveebg}\textbf{64.07}\posdelta{24.32} \\
        \bottomrule
    \end{tabular}%
    }
\end{table*}

\subsection{Ablations}

Table~\ref{tab:ablation-results} isolates the main components on Qwen3-4B-Instruct. Besides the unadapted baseline, we test a default router that replaces the learned routing field with a blank or simple default field, a manual router written once by GPT-5.4 and then fixed, and a no-co-evolution variant that first learns the router and then learns prompts in a separate second stage.

The full method reaches 51.75 average score, 8.17 points above the default router (43.58), 14.57 above the manual router (37.18), and 8.87 above no co-evolution (42.88). Default routing and no co-evolution improve the baseline by only 2.21 and 1.51 points, respectively, while the manual router is 4.19 points below the baseline. These gaps show that \textsc{Eevee} needs both learned routing and interleaved router--prompt optimization; static partitions or two-stage learning do not capture the mutual dependence between routing decisions and prompt behavior.

\begin{table*}[htbp]
    \centering
    \caption{Ablation results for the main components of \textsc{Eevee} on Qwen3-4B-Instruct.}
    \label{tab:ablation-results}
    \vspace{0.1cm}
    \resizebox{0.8\textwidth}{!}{%
    \begin{tabular}{l|cccc|c}
        \toprule
        Variant & GPQA Diamond & Formula & TheoremQA & HumanEval & Avg. \\
        \midrule
        Baseline & 56.00 & 45.22 & 14.79 & 49.46 & 41.37 \\
        Default Router & 47.03 & 47.11 & 26.10 & 54.06 & 43.58 \\
        Manual Router & 35.24 & 42.11 & 23.96 & 47.42 & 37.18 \\
        No Co-evolution & 39.84 & 47.39 & 24.81 & 59.49 & 42.88 \\
        \rowcolor{eeveebg} \textsc{Eevee} & 54.55 & 54.55 & 25.27 & 72.63 & 51.75 \\
        \bottomrule
    \end{tabular}%
    }
\end{table*}

\subsection{Task Scaling}

\begin{figure*}[htbp]
    \centering
    \includegraphics[width=\linewidth]{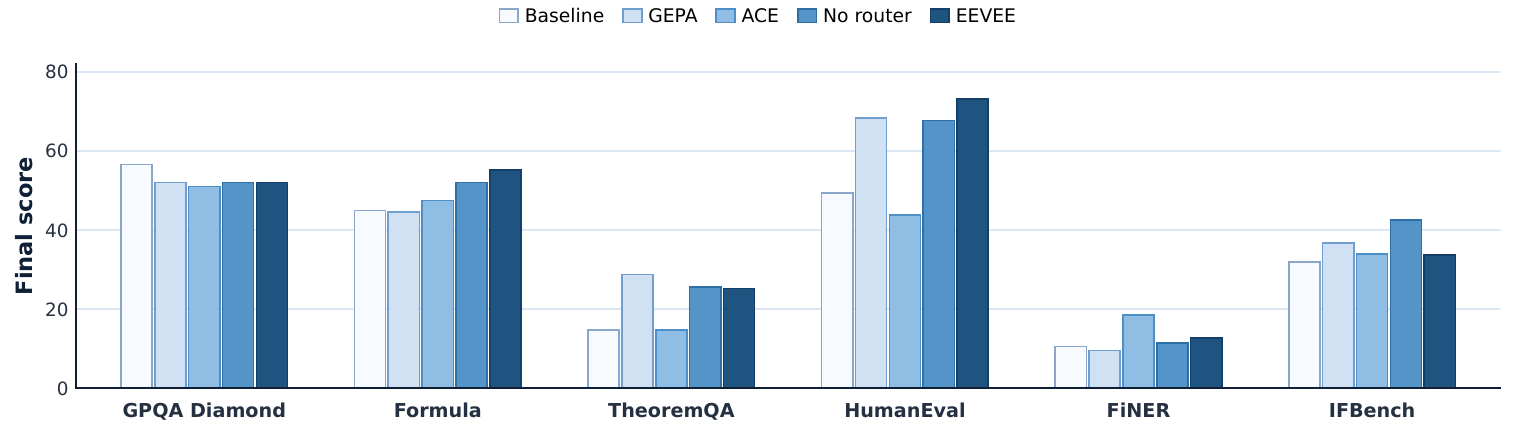}
    \caption{Single-benchmark results~\cite{loukas2022finer,pyatkin2025ifbench}. Bars show final scores after learning on each benchmark independently. The no-router variant keeps a single retained prompt and therefore isolates prompt learning without router specialization.}
    \label{fig:single-benchmark}
\end{figure*}

We first check whether \textsc{Eevee} remains competitive when prompt learning is run on one benchmark, then study what happens as the number of jointly learned benchmarks increases. In the single-benchmark setting in Figure~\ref{fig:single-benchmark}, \textsc{Eevee} is broadly competitive with GEPA and ACE. Figure~\ref{fig:single-benchmark} also includes FiNER~\cite{loukas2022finer} and IFBench~\cite{pyatkin2025ifbench} to align with benchmarks tested by prior methods. It reaches 55.25 on Formula and 73.17 on HumanEval, outperforming both baselines on these two benchmarks, and improves TheoremQA from 14.73 to 25.27. The no-router variant also performs strongly on Formula and HumanEval, showing that our prompt learning design itself is effective even without router specialization.

The difference becomes clearer as the benchmark mixture grows. Figure~\ref{fig:multi-benchmark-curve} shows that GEPA and ACE quickly lose retention as more tasks are added; after all four tasks, both end below zero. In contrast, \textsc{Eevee} remains positive throughout and ends at +41.53 cumulative retention. Thus the main advantage appears when the task mixture grows: a single learned prompt loses retention, while router-conditioned prompt learning preserves positive net improvement.

\subsection{Generalization}

We test two forms of generalization: cross-model generalization, which asks whether prompts learned on one target model still help a different target model, and cross-task generalization, which asks whether prompts learned on the four primary benchmarks generalize to unseen tasks. For cross-model generalization, prompts learned with Qwen3-4B-Instruct~\cite{yang2025qwen3} are applied directly to DeepSeek-V3.2~\cite{deepseekai2025v32} in non-thinking mode. For cross-task generalization, prompts learned on the four primary benchmarks are evaluated on held-out MBPP~\cite{austin2021program} and MMLU-Pro~\cite{wang2024mmlupro}. MBPP is close to HumanEval and tests coding-domain generalization; MMLU-Pro is broader knowledge QA and tests unrelated-domain robustness.

\begin{table*}[htbp]
    \centering
    \small
    \caption{Cross-model and held-out benchmark generalization. Left: prompts learned with Qwen3-4B-Instruct are directly evaluated on DeepSeek-V3.2, alongside the source-model learned result. Right: prompts learned on the main four-benchmark suite are evaluated on MBPP and MMLU-Pro, using a shared unadapted baseline. Scores are averaged over three runs.}
    \vspace{0.1cm}
    \label{tab:generalization}
    \setlength{\tabcolsep}{4pt}
    \renewcommand{\arraystretch}{1.08}
    \begin{minipage}[t]{0.67\textwidth}
        \centering
        \textbf{Cross-Model Generalization}\\[2pt]
        \resizebox{\linewidth}{!}{%
        \begin{tabular}{l|cccc|c}
            \toprule
            Setting & GPQA Diamond & Formula & TheoremQA & HumanEval & Avg. \\
            \midrule
            DeepSeek baseline & 64.98 & 30.00 & 21.21 & 42.82 & 39.75 \\
            Qwen3 \textsc{Eevee} & 54.55 & 54.55 & 25.27 & 72.63 & 51.75 \\
            \rowcolor{eeveebg} Qwen3$\rightarrow$DeepSeek & 60.27 & 42.28 & 32.89 & 77.04 & \textbf{54.10} \\
            \bottomrule
        \end{tabular}
        }
    \end{minipage}
    \hfill
    \begin{minipage}[t]{0.3\textwidth}
        \centering
        \textbf{Held-Out Generalization}\\[2pt]
        \resizebox{\linewidth}{!}{%
        \begin{tabular}{l|cc|c}
            \toprule
            Method & MBPP & MMLU-Pro & Avg. \\
            \midrule
            Baseline & 69.29 & 70.74 & 70.01 \\
            \rowcolor{eeveebg} \textsc{Eevee} & \textbf{70.42} & 68.92 & 69.67 \\
            GEPA & 68.20 & 68.85 & 68.53 \\
            ACE & 67.47 & 69.32 & 68.39 \\
            \bottomrule
        \end{tabular}
        }
    \end{minipage}
\end{table*}

Table~\ref{tab:generalization} summarizes both settings. In cross-model generalization, the prompts learned on Qwen3-4B-Instruct raise the DeepSeek-V3.2 average from 39.75 to 54.10, with gains of +12.28 on Formula, +11.68 on TheoremQA, and +34.22 on HumanEval. In cross-task generalization, \textsc{Eevee} improves MBPP from 69.29 to 70.42, while GEPA and ACE drop to 68.20 and 67.47. On MMLU-Pro, \textsc{Eevee} decreases from 70.74 to 68.92, a 1.82-point drop that is smaller than GEPA's 1.89-point drop and comparable to ACE's 1.42-point drop. This pattern is aligned with our later case study, where we further analyze its underlying causes.

\subsection{Token Cost}

\begin{figure}[htbp]
    \centering
    \includegraphics[width=0.88\linewidth]{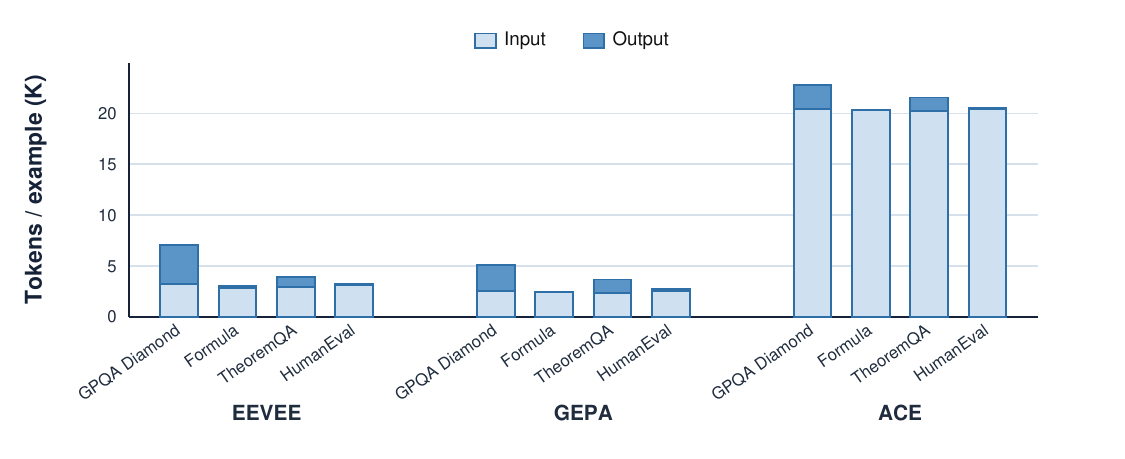}
    \caption{Average token usage per test example after test-time prompt learning with Qwen3-4B-Instruct. Each method group contains four benchmark bars, and each bar is stacked into input and output tokens.}
    \label{fig:token-cost}
\end{figure}

\textsc{Eevee} adds a router before answer generation, so we measure final-test token cost. Averaged over the four benchmarks in Figure~\ref{fig:token-cost}, \textsc{Eevee} uses 4.32k total tokens per example, close to GEPA's 3.47k and far below ACE's 21.30k. The input-token gap is similar: \textsc{Eevee} uses 3.00k input tokens on average, compared with 2.44k for GEPA and 20.35k for ACE. ACE incrementally edits playbook bullets, which can accumulate redundant entries and lengthen prompts as tasks and data grow. \textsc{Eevee}'s router therefore adds only modest overhead while using about 4.9$\times$ fewer total tokens than ACE.

\subsection{Case Study: What Does Prompt Learning Capture?}
\label{sec:case-study}

We retest six completed \textsc{Eevee} runs by comparing the empty prompt with the final learned router and prompt set on the same held-out examples: three runs with Qwen3-4B-Instruct and three runs with DeepSeek-V3.2. Table~\ref{tab:case-study-retest} reports the flip summary; the cases below show what these flips reveal, with representative learned prompt and output excerpts in Appendix~\ref{app:case-study-details}.

\begin{table*}[htbp]
    \centering
    \small
    \caption{Six-run diagnostic empty-vs-final retest over three Qwen3-4B-Instruct and three DeepSeek-V3.2 runs. Qwen $\Delta$ and DeepSeek $\Delta$ are average percentage-point final-minus-empty changes within each model family. W$\rightarrow$R/R$\rightarrow$W aggregates all per-example correctness flips across the six retests. Runs+ counts runs with positive $\Delta$.}
    \vspace{0.1cm}
    \label{tab:case-study-retest}
    \setlength{\tabcolsep}{6pt}
    \renewcommand{\arraystretch}{1.08}
    \begin{tabular}{llcccc}
        \toprule
        Type & Benchmark & Qwen $\Delta$ & DeepSeek $\Delta$ & W$\rightarrow$R/R$\rightarrow$W & Runs+ \\
        \midrule
        Code & HumanEval & \textcolor{gaincolor}{+23.2} & \textcolor{gaincolor}{+48.8} & 193 / 16 & 6/6 \\
        Formula & Formula & \textcolor{gaincolor}{+9.5} & \textcolor{gaincolor}{+31.7} & 268 / 21 & 6/6 \\
        Math QA & TheoremQA & \textcolor{gaincolor}{+10.0} & \textcolor{gaincolor}{+18.2} & 632 / 166 & 6/6 \\
        Domain QA & GPQA Diamond & \textcolor{dropcolor}{-3.7} & \textcolor{dropcolor}{-7.7} & 55 / 89 & 1/6 \\
        \midrule
        Overall & Mixed suite & \textcolor{gaincolor}{+9.6} & \textcolor{gaincolor}{+21.1} & 1148 / 292 & 6/6 \\
        \bottomrule
    \end{tabular}
\end{table*}

The table suggests a task-property pattern rather than a benchmark-specific one. Gains are larger on code and formula tasks, where feedback can be turned into reusable rules for how to solve and how to present the answer. TheoremQA also improves, but with more mixed flips because it combines computation, symbolic reasoning, and answer parsing. GPQA Diamond is the only benchmark with more regressions than recoveries, suggesting that stronger learned reasoning can sometimes underweight the domain knowledge needed for closed-book QA. Figure~\ref{fig:case-study-cases} shows two positive flips and one representative negative flip.

\begin{figure*}[htbp]
    \centering
    \setlength{\fboxsep}{5pt}
    \fcolorbox{casegreenborder}{casegreenbg}{%
    \begin{minipage}[t]{0.285\textwidth}
        \small\raggedright
        \textbf{Formula: unit scale}\\[2pt]
        \textbf{Task}\\
        Compute free cash flow from operating cash flow and capital expenditure.\\[3pt]
        \textbf{Baseline}\\
        Reverses the subtraction and keeps the million-scale decimal, producing a negative value.\\[3pt]
        \textbf{Learned}\\
        Applies the formula in dollars and emits the strict numeric answer: 400000.00.
    \end{minipage}}
    \hfill
    \fcolorbox{caseblueborder}{casebluebg}{%
    \begin{minipage}[t]{0.285\textwidth}
        \small\raggedright
        \textbf{HumanEval: executable body}\\[2pt]
        \textbf{Task}\\
        Complete a function that sums even values appearing at odd indices.\\[3pt]
        \textbf{Baseline}\\
        Writes a bare expression without the required indented return statement.\\[3pt]
        \textbf{Learned}\\
        Produces an executable function body with an accumulator and return.
    \end{minipage}}
    \hfill
    \fcolorbox{caseorangeborder}{caseorangebg}{%
    \begin{minipage}[t]{0.285\textwidth}
        \small\raggedright
        \textbf{GPQA Diamond: knowledge underweighted}\\[2pt]
        \textbf{Task}\\
        Select the densest Earth-like exoplanet from mass and composition cues.\\[3pt]
        \textbf{Baseline}\\
        Uses the rocky-planet prior that higher mass increases self-compression and density.\\[3pt]
        \textbf{Learned}\\
        Performs formulaic density reasoning, treats same composition as constant density, and selects the Earth baseline.
    \end{minipage}}
    \caption{Representative case-study flips. The boxes summarize task behavior rather than reproducing raw inputs; representative learned prompt and output excerpts are provided in Appendix~\ref{app:case-study-details}.}
    \label{fig:case-study-cases}
\end{figure*}

These cases explain why prompt learning can both improve and hurt. For executable programs and formula-grounded computation, feedback teaches a better procedure: preserve the interface, handle edge cases, respect units, and emit a parseable final answer. For knowledge-intensive QA, the learned prompt often strengthens generic reasoning and answer-comparison behavior rather than adding the specific missing knowledge. In the GPQA Diamond example, the learned response reasons more explicitly from density formulas, but it relies on a constant-density assumption and underweights the astrophysical fact that rocky planets become denser under gravitational compression.

\vspace{0.5cm}
\noindent\makebox[\linewidth][c]{\fcolorbox{caseblueborder}{casebluebg}{%
\begin{minipage}[c][2.5\baselineskip][c]{0.965\linewidth}
    \small\centering
    \textbf{Takeaway.} Prompt learning excels at reusable procedures, but can underweight domain knowledge.
\end{minipage}}}

\section{Related Work}

\paragraph{Prompt learning.}
Prompt learning first optimized soft prompts, prefixes, or discrete triggers for fixed objectives~\cite{shin2020autoprompt,li2021prefix,lester2021power}; black-box and population-based methods later use the model, scores, or textual feedback to optimize prompts and programs~\cite{zhou2022large,yang2024large,pryzant2023automatic,guo2024connecting,fernando2023promptbreeder,khattab2023dspy,opsahlong2024optimizing,yuksekgonul2024textgrad}. Recent reflective methods are closest to our work: GEPA~\cite{agrawal2026gepa} uses natural-language reflection and Pareto-front selection~\cite{mouret2015illuminating}, ACE treats context as an adaptive playbook~\cite{zhang2026agentic}, and Combee scales prompt learning with parallel trace aggregation~\cite{li2026combee}. These methods improve feedback-driven adaptation, but typically optimize one task distribution or one shared prompt/context. Recent heterogeneous memory extraction work also learns prompts across many datasets, but its target remains memory extraction rather than general LLM-agent capabilities~\cite{yang2026selfevolvingmemory}. \textsc{Eevee} instead learns a router-conditioned prompt set for task-general, mixed-dataset streams.

\paragraph{Self-improving agents.}
Self-improving agents extend prompt learning into feedback loops. Self-Refine and Reflexion use natural-language feedback or verbal memory~\cite{madaan2023selfrefine,shinn2023reflexion}, while generative agents and Voyager maintain longer-lived memories or skill libraries~\cite{park2023generative,wang2023voyager}. Evolutionary discovery agents apply similar loops to scientific and algorithmic search, including code-candidate evolution and adaptive search control~\cite{novikov2025alphaevolve,sharma2025openevolve,liu2026skydiscover,cemri2026adaevolve,liu2026evox,xu2026asievolve}. These systems show that histories can drive improvement, but they mainly optimize scoped programs or algorithms. \textsc{Eevee} instead targets real-world heterogeneous tasks, where effective improvement requires decoupling multi-task commonality from task-specific behavior.

\begin{table}[t]
    \centering
    \small
    \caption{Comparison on mixed-dataset adaptation, router-based prompt selection, and joint router-prompt co-evolution.}
    \label{tab:related-work-comparison}
    \begin{tabular}{lccc}
        \toprule
        Method & Multi-dataset & \begin{tabular}{@{}c@{}}Router\\(Sec.~\ref{sec:framework-overview})\end{tabular} & \begin{tabular}{@{}c@{}}Co-evolution\\(Sec.~\ref{sec:co-evolution})\end{tabular} \\
        \midrule
        Prompt tuning/search & $\times$ & $\times$ & $\times$ \\
        GEPA / ACE / Combee & $\times$ & $\times$ & $\times$ \\
        Reflexive agents & $\times$ & $\times$ & $\times$ \\
        Evolutionary discovery agents & $\times$ & $\times$ & \checkmark \\
        \midrule
        \rowcolor{eeveebg} \textsc{Eevee} & \checkmark & \checkmark & \checkmark \\
        \bottomrule
    \end{tabular}
\end{table}

\section{Conclusion}

We introduced \textsc{Eevee}, a multi-dataset test-time prompt learning framework for LLM agents facing heterogeneous task streams rather than a single benchmark distribution. To reduce cross-dataset interference, \textsc{Eevee} maintains a router-conditioned prompt set, so different inputs can be assigned to prompts specialized for compatible task behavior. Because the router and prompts depend on each other, \textsc{Eevee} uses a three-stage router-prompt co-evolution procedure that initializes useful prompts, explores coupled updates, and then refines prompts under a stable router. This makes the framework better aligned with heterogeneous real-world agent workloads.

Experiments show strong gains over prompt-learning baselines in mixed-dataset settings, with benefits increasing as more tasks are introduced and with reasonable held-out and cross-model transfer. Case studies further suggest that prompt learning is most useful when feedback can be converted into reusable procedures, output contracts, or task-solving strategies. Overall, \textsc{Eevee} offers both a practical self-improving method and an empirical lens for real-world test-time prompt learning.

\section{Limitations and Social Impact}
\label{sec:limitations-social-impact}

Although \textsc{Eevee} improves multi-dataset prompt learning, limitations remain. Like other LLM-based evolutionary procedures, it cannot guarantee exact performance reproduction across runs, since stochastic search can produce different routers and prompt sets. Its feedback loop still relies on ground-truth or rule-based labels to accumulate task knowledge, so it is not yet a fully reflection-only learner and still needs a prepared adaptation set rather than a completely online stream. A practical risk is distribution mismatch: if adaptation data is noisy or misaligned with the real application, learned prompts may generalize weakly or even degrade performance. Deployment should check data quality and distribution similarity before updates.

\bibliographystyle{plainnat}
\bibliography{reference}

\appendix

\section{Case Study Details}
\label{app:case-study-details}

This appendix provides excerpts from the diagnostic retest discussed in Section~\ref{sec:case-study}. The retest compares the empty prompt against the final router and prompt set from six completed \textsc{Eevee} runs: three Qwen3-4B-Instruct runs and three DeepSeek-V3.2 runs. The raw run logs store the full router decisions, answer calls, and per-example flip bundles. We do not reproduce all six prompt sets; instead, we include representative learned prompt excerpts from one Qwen3-4B-Instruct run and one DeepSeek-V3.2 run, followed by output excerpts for the three main-paper cases.

\subsection{Learned Prompt Excerpts}

\paragraph{Qwen3-4B code slot.}
In a representative Qwen3-4B-Instruct run, the code-oriented slot learned a task-execution policy for incomplete Python functions: preserve the interface, infer the exact continuation, cover edge cases, and avoid extra text that would break execution.

\begin{verbatim}
You are given a task description that defines an incomplete Python
function or problem to solve. Carefully read the input, extract
the exact task requirements, and generate only the precise output
expected by the task without extra text, explanations, markdown,
comments, or formatting.

Identify the task type:
- Writing a Python function body with specific inputs/outputs
- Solving a mathematical or logical problem
- Validating conditions such as bracket matching or palindrome checks
- Processing structured data such as strings, arrays, and dictionaries

Parse the input format and constraints:
- Note data types, edge cases, boundary values, and examples
- Pay close attention to ordering and special constraints

Understand the output contract:
- If the task says output ONLY the continuation,
  write only the function body.
\end{verbatim}

\paragraph{Qwen3-4B science slot.}
The same representative Qwen3-4B-Instruct run routes many GPQA Diamond examples to a scientific-reasoning slot. This prompt encourages systematic physical and mathematical modeling, but it does not provide task-specific missing knowledge for every possible GPQA Diamond domain.

\begin{verbatim}
You are an expert problem solver specializing in analytical reasoning,
physical modeling, combinatorial mathematics, and scientific computation.
Your task is to answer precise, domain-specific questions that require
careful application of physical laws, mathematical principles, or
combinatorial logic, based on exact input constraints and known scientific
or mathematical facts.

For multiple-choice questions with options (A-D), select the single correct
choice and present it in the format: "Answer: (X)".

For physical or astronomical problems involving equilibrium, energy,
temperature, or orbital dynamics, apply known physical laws such as the
Stefan-Boltzmann law, Kepler's laws, Doppler shift formula, blackbody
radiation, and Newtonian gravity.
\end{verbatim}

\paragraph{DeepSeek-V3.2 formula slot.}
In a representative DeepSeek-V3.2 run, the formula slot learned a stricter numerical-output policy. This slot is especially aligned with the Formula benchmark, where the input gives an explicit formula and the evaluator expects a compact numeric answer.

\begin{verbatim}
You are a finance calculation assistant. Your task is to compute the
answer using the formula and data provided in the user's question,
then output only the numeric result with exactly two decimal places
and no other text.

Procedure:
1. Identify the formula given in the user's message.
2. Identify the explanation of each variable in the formula.
3. Extract the numeric values for each variable from the question text.
   Convert all percentage inputs into their decimal equivalent.
4. Substitute the values into the formula and perform the calculation.
5. If the formula calculates a financial rate, yield, return, or cost,
   do not multiply the decimal result by 100.
6. Round the final result to two decimal places.
7. Output only the resulting number, with no words, units, labels,
   currency symbols, or percentage signs.
\end{verbatim}

\subsection{Representative Raw Outputs}

\paragraph{Formula: unit scale and sign.}
For a free-cash-flow computation, the required operation is operating cash flow minus capital expenditure. The target is a dollar-scale scalar value. The empty response flips the sign and keeps the million-scale decimal:

\begin{verbatim}
-0.40
\end{verbatim}

The learned response applies the provided formula at the correct scale and emits the strict numeric answer:

\begin{verbatim}
Answer: 400000.00
\end{verbatim}

This is representative of the Formula gains: the learned prompt pairs formula application with unit-scale discipline and a parseable final representation.

\paragraph{HumanEval: executable continuation.}
For a HumanEval function-completion task, the required behavior is to sum even elements at odd indices and return the sum. The empty response captures the expression but does not produce an executable function body:

\begin{verbatim}
sum(lst[i] for i in range(1, len(lst), 2) if lst[i] % 2 == 0)
\end{verbatim}

The learned response preserves the logic while satisfying the continuation contract:

\begin{verbatim}
    total = 0
    for i in range(1, len(lst), 2):
        if lst[i] % 2 == 0:
            total += lst[i]
    return total
\end{verbatim}

This is representative of the HumanEval gains: the learned behavior combines implementation logic with the executable contract expected by the evaluator.

\paragraph{GPQA Diamond: domain knowledge underweighted after stronger reasoning.}
For an exoplanet-density question, the target is the higher-mass Earth-composition planet. The empty response uses a domain-specific mass-radius prior:

\begin{verbatim}
For rocky planets of similar composition, radius grows sublinearly with
mass because stronger gravity compresses the material. A five-Earth-mass
rocky planet is therefore denser than Earth, while a half-Earth-mass
rocky planet is less dense.

Answer: the higher-mass Earth-composition option.
\end{verbatim}

The learned response performs a more explicit density-formula comparison, but it treats equal composition as if it implied equal density:

\begin{verbatim}
Density is mass divided by volume. For a spherical planet,
rho = M / (4/3 pi R^3). If composition is the same, density is
constant; the radius scales with mass so that all Earth-composition
options remain approximately Earth density. Since the Earth-mass,
Earth-radius option is exactly Earth-like, choose that option.
\end{verbatim}

The error is not a lack of reasoning steps. The learned prompt induces a stronger generic reasoning pattern, but the reasoning is grounded in the wrong physical prior: same composition does not imply constant density for rocky exoplanets, because self-compression changes the mass-radius relation.

\section{Reproducibility and Experimental Details}
\label{app:reproducibility}

This section records the core settings used for the main four-benchmark experiments in Table~\ref{tab:main-results}. The benchmark suite consists of GPQA Diamond~\cite{rein2023gpqa}, Formula~\cite{wang2025finlora}, TheoremQA~\cite{chen2023theoremqa}, and HumanEval~\cite{chen2021codex}. The held-out generalization study uses MBPP~\cite{austin2021program} and MMLU-Pro~\cite{wang2024mmlupro} after prompt learning on the four primary benchmarks. The single-benchmark diagnostic also includes FiNER~\cite{loukas2022finer} and IFBench~\cite{pyatkin2025ifbench}. For each benchmark, we cap the benchmark size at 500 examples, split examples into train and test partitions with a $0.5/0.5$ split, reserve half of the training partition as validation data, exclude train and validation examples from the final test set, and use data seed 42.

The main target models are Qwen3-4B-Instruct~\cite{yang2025qwen3} and DeepSeek-V3.2~\cite{deepseekai2025v32} in non-thinking mode. For the Qwen3-4B-Instruct runs, all model roles use the same Qwen endpoint with temperature 0.7, top-$p$ 0.8, and a maximum generation length of 16,384 tokens. For the DeepSeek-V3.2 non-thinking runs, the endpoint uses temperature 1.0, top-$p$ 0.95, maximum generation length 8,192 tokens, and disabled thinking. The main benchmark configurations use rule-based benchmark judges when available.

\paragraph{Evolution settings.}
Each main run retains four bootstrap prompts. Bootstrap prompt evolution uses a budget of 10 candidate steps. The router--prompt co-evolution phase uses a total mini-step budget of 150, router and prompt windows of 3, and phase-switch threshold 0.005. The router score during evolution combines downstream score, routing consistency, and balance with weights $0.6/0.2/0.2$; the final router selection uses downstream score only, with weights $1.0/0.0/0.0$. Prompt evolution uses a final per-label budget of 60, minibatch size 5, and 4 parallel prompt slots.

\paragraph{Execution settings and resources.}
The Qwen3-4B-Instruct main runs use 50 API workers for evaluation, router calls, and final testing. The DeepSeek-V3.2 non-thinking main runs use 10 API workers for the same roles. All reported final tests use three repeats, a maximum prompt length of 2,000 tokens, at most 3 retries, and at most 5 logged examples for diagnostic traces. The experiments call hosted or served API endpoints and do not train or fine-tune local model weights; accordingly, no local GPU training resource is required. Wall-clock time is backend- and rate-limit-dependent, so the paper reports token usage instead. In Figure~\ref{fig:token-cost}, \textsc{Eevee} uses 4.32k total tokens per test example on average, compared with 3.47k for GEPA and 21.30k for ACE.

\paragraph{Reproducibility scope.}
The paper is empirical and does not present formal theoretical results or proofs. Reproducing the main claims requires reproducing the mixed-benchmark adaptation protocol, model endpoints or comparable served checkpoints, benchmark splits, router--prompt evolution settings, and final-test evaluation described above. Because router and prompt evolution are stochastic, exact routers and prompt texts may vary across runs even under the same settings; Table~\ref{tab:main-average-std} reports this run-to-run variation for the main average score.

\subsection{Hyperparameter Robustness}
\label{app:hyperparameter-robustness}

We test whether \textsc{Eevee} is sensitive to the specific router-score and prompt-search hyperparameters used in the main experiments. For each hyperparameter configuration, we run three independent trials, first average the three trials within that configuration, and then compare the resulting eight configuration-level means. The configurations vary the annealing target for router scores, the consistency/balance weights, and the final prompt-search budget, minibatch size, and temporary prompt-pool size.

\begin{table*}[htbp]
    \centering
    \small
    \caption{Hyperparameter robustness on Qwen3-4B-Instruct. Scores are percentages; each row is the mean over three independent runs for one configuration. Avg. is the macro average over the four benchmarks.}
    \label{tab:hyperparameter-robustness}
    \vspace{0.1cm}
    \setlength{\tabcolsep}{5pt}
    \renewcommand{\arraystretch}{1.05}
    \resizebox{0.92\textwidth}{!}{%
    \begin{tabular}{l|ccccc}
        \toprule
        Configuration & GPQA Diamond & Formula & TheoremQA & HumanEval & Avg. \\
        \midrule
        No-decay annealing & 49.94 & 52.17 & 25.66 & 58.67 & 46.61 \\
        Soft-landing annealing & 45.01 & 57.44 & 21.33 & 69.24 & 48.26 \\
        Strong-regularizer annealing & 45.34 & 46.61 & 24.57 & 63.69 & 45.05 \\
        Classic reference & 44.78 & 60.06 & 26.20 & 57.99 & 47.26 \\
        Prompt search tighter & 47.92 & 53.33 & 27.41 & 63.01 & 47.92 \\
        Prompt search wider & 48.15 & 62.39 & 23.94 & 61.38 & 48.96 \\
        Consistency-heavy weights & 51.74 & 50.33 & 26.79 & 63.96 & 48.20 \\
        Balance-heavy weights & 49.16 & 54.89 & 26.93 & 72.90 & 50.97 \\
        \bottomrule
    \end{tabular}%
    }
\end{table*}

The eight configuration-level averages range from 45.05 to 50.97, a span of 5.92 points, with a sample standard deviation of 1.73 points across configurations. Thus the aggregate result is stable under the tested hyperparameter perturbations. Individual benchmarks can move more than the average, but no configuration collapse is observed, and every configuration improves over its corresponding initial-empty baseline in macro average.

\subsection{Main-Result Variation Across Runs}
\label{app:run-variation}

Table~\ref{tab:main-average-std} reports the mean and sample standard deviation of the main average score over the three independent runs used in Table~\ref{tab:main-results}. The average score is stable for \textsc{Eevee}: the standard deviation is 1.62 points on Qwen3-4B-Instruct and 1.08 points on DeepSeek-V3.2. Individual benchmark scores can still differ more noticeably across runs. The reason is that router evolution is stochastic and can discover different routing policies; different policies allocate examples to different prompt slots, which changes which prompt behaviors receive the most feedback and can shift per-task scores even when the overall average remains stable.

\begin{table}[htbp]
    \centering
    \small
    \caption{Mean and sample standard deviation of the main average score over three runs.}
    \label{tab:main-average-std}
    \begin{tabular}{llc}
        \toprule
        Target model & Method & Average score \\
        \midrule
        \multirow{4}{*}{Qwen3-4B-Instruct}
        & Baseline & $41.37 \pm 0.51$ \\
        & ACE & $34.92 \pm 4.14$ \\
        & GEPA & $37.73 \pm 1.73$ \\
        & \textsc{Eevee} & $\mathbf{51.75 \pm 1.62}$ \\
        \midrule
        \multirow{4}{*}{DeepSeek-V3.2}
        & Baseline & $39.75 \pm 1.52$ \\
        & ACE & $49.83 \pm 2.07$ \\
        & GEPA & $55.83 \pm 4.48$ \\
        & \textsc{Eevee} & $\mathbf{64.07 \pm 1.08}$ \\
        \bottomrule
    \end{tabular}
\end{table}

\section{Ethics, Assets, and LLM Usage}
\label{app:ethics-release}

\paragraph{Human subjects and privacy.}
The experiments use public benchmarks and model API calls. We do not collect new human-subject data, run crowdsourcing studies, or introduce a dataset containing personal information.

\paragraph{Existing assets and code availability.}
The experiments use public benchmarks, public or provider-served model checkpoints, and published prompt-learning baselines such as GEPA and ACE. Code, configuration files, reproduction scripts, and asset metadata are released in the official repository at \url{\codeurl}. The accompanying project page is available at \url{\projectpageurl}. This work introduces a method rather than a new dataset or model checkpoint.

\paragraph{Responsible use and broader impacts.}
\textsc{Eevee} can make heterogeneous-task prompt adaptation more efficient by reducing the need to maintain one prompt-learning run per task family. The main practical risk is that noisy, incomplete, or distribution-shifted feedback can reinforce incorrect heuristics. Adapted prompts should be validated on held-out data before deployment, and benchmark gains should not be interpreted as deployment reliability guarantees.

\paragraph{LLM usage.}
LLMs are core components of the method: they act as the target model being adapted and as the prompt researcher, prompt reflector, router selector, router researcher, router reflector, router reasoner, and evaluator where applicable. In the ablation study, GPT-5.4 was used once to write a fixed manual router, which was then held constant for evaluation. LLMs were also used for language editing and formatting of the manuscript; this editing use did not change the scientific claims, experimental data, or conclusions.

\end{document}